\documentclass[conference]{IEEEtran}

\ifCLASSINFOpdf
  \usepackage[pdftex]{graphicx}
\else
\fi

\usepackage{float}
\usepackage{cite}

\begin{document}

\title{Dynamic LIBRAS Gesture Recognition via CNN over Spatiotemporal Matrix Representation}

\author{\IEEEauthorblockN{Jasmine Moreira}
\IEEEauthorblockA{CPGEI\\
Universidade Tecnológica Federal do Paraná\\
Curitiba, Paraná, Brazil\\
Email: jasmine.moreira.2013@gmail.com}}

\maketitle

\begin{abstract}
This paper proposes a method for dynamic hand gesture recognition based on the composition of two models: the MediaPipe Hand Landmarker, responsible for extracting 21 skeletal keypoints of the hand, and a convolutional neural network (CNN) trained to classify gestures from a spatiotemporal matrix representation of dimensions 90 by 21 of those keypoints. The method is applied to the recognition of LIBRAS (Brazilian Sign Language) gestures for device control in a home automation system, covering 11 classes of static and dynamic gestures. For real-time inference, a sliding window with temporal frame triplication is used, enabling continuous recognition without recurrent networks. Tests achieved 95\% accuracy under low-light conditions and 92\% under normal lighting. The results indicate that the approach is effective, although systematic experiments with greater user diversity are needed for a more thorough evaluation of generalization.
\end{abstract}


\section{Introduction}
Automatic hand gesture recognition is a fundamental component of vision-based human-machine interfaces, with applications ranging from device control to sign language interpretation. Despite recent advances in deep learning, the precise recognition of \textit{dynamic} gestures — those involving movement over time — remains a challenge. Hand anatomy is particularly complex, with joints, tendons, and proportions that vary across individuals, making it difficult to build models that generalize well to multiple users.

Representing this variability requires capturing both the spatial configuration of the hand at each instant and the evolution of that configuration over time. Approaches based on raw image pixels impose high dimensionality and are sensitive to factors such as lighting and skin color. A more robust alternative is the use of skeletal keypoints extracted by specialized models, which abstract the visual appearance and retain only the geometric structure relevant to gesture recognition.

Several individual and collective initiatives address this problem. Many of them rely on well-established computer vision libraries such as OpenCV\footnote{https://opencv.org} and YOLO\footnote{https://docs.ultralytics.com}. One of the most productive initiatives is the \textit{Hand Landmarker}\footnote{http://developers.google.com/mediapipe/solutions/vision/hand\_landmarker}, from the MediaPipe group, associated with Google.

Through this tool, it is possible to identify static hand positions in an image. This allows different strategies to be used for dynamic gesture recognition, captured in videos or image sequences. Among the most common strategies are comparisons with predefined poses, performed through mathematical, statistical, or neural network operations.

However, the recognition task becomes more difficult when dealing with dynamic gestures, whose execution involves hand movements of different forms and at different time intervals. One example is the letter J in the LIBRAS alphabet, which involves a specific finger position associated with hand rotation.

This paper proposes a dynamic gesture recognition technique aimed at tolerating variations in expression rhythm and finger and hand positioning, with application to home automation device control through LIBRAS gestures. The main contributions of this work are: (1) a spatiotemporal matrix representation (90$\times$21) of skeletal keypoints extracted by MediaPipe, organized by axis and frame, treated as an image for classification by a 2D CNN; (2) a sliding window strategy with temporal frame triplication for continuous real-time inference, without recurrent networks; and (3) validation of the system in a home automation scenario with static and dynamic LIBRAS gestures covering 11 classes.

\section{Related Work}

Dynamic hand gesture recognition has been widely investigated in the literature, with approaches ranging from recurrent networks to three-dimensional convolutional networks. This section contextualizes the present work in relation to the most relevant recent studies.

In the context of LIBRAS recognition with deep learning, Rezende et al. \cite{IEEEhowto:rezende} introduced the MINDS-Libras dataset, comprising 1,200 videos of 20 signs performed by 12 users, achieving 93.3\% accuracy with the best evaluated classifier. Amaral et al. \cite{IEEEhowto:amaral} evaluated deep models for dynamic LIBRAS recognition from depth data, with LRCN and 3D CNN networks achieving above 99\% under controlled conditions. More recently, Alves et al. \cite{IEEEhowto:alves} proposed an approach based on skeletal image representation, encoding MediaPipe-extracted keypoints as spatiotemporal images classified by CNN, obtaining 93\% on MINDS-Libras and 82\% on LIBRAS-UFOP with multiple users. This approach is conceptually close to the one proposed in this paper, differing mainly in the organization of the keypoint matrix and in the use of a sliding window with temporal frame triplication.

Regarding MediaPipe-based pipelines for dynamic recognition, Yaseen et al. \cite{IEEEhowto:yaseen} combined MediaPipe keypoint extraction with Inception-v3 and LSTM, achieving 89.7\% on the 20BN-Jester dataset. Ridwang et al. \cite{IEEEhowto:ridwang} used MediaPipe with a modified LSTM for dynamic sign language recognition, reporting 99.4\% training accuracy and 85\% per-word accuracy in real time, highlighting the gap between training accuracy and production performance. Hakim et al. \cite{IEEEhowto:hakim} proposed a combination of 3D CNN with LSTM and a finite state machine for context control in continuous gesture recognition, achieving 97.6\% on isolated tests and 93\% on real-time sequences.

In the area of gesture-based home automation, Alabdullah et al. \cite{IEEEhowto:alabdullah} proposed a domestic control system with feature fusion and RNN, achieving 92.57\% on the HaGRI dataset. Compared to these works, the approach proposed in this paper adopts a lightweight 2D CNN architecture ($\sim$25,000 parameters) over the (90$\times$21) skeletal keypoint matrix, dispensing with recurrent networks. The temporal triplication of frames in the inference buffer and the 98\% confidence threshold contribute to real-time robustness on general-purpose hardware.

\section{Methodology}

The method for dynamic gesture detection proposed in this paper is based on a composition of two models. The first is the MediaPipe Hand Landmarker \cite{IEEEhowto:zhang}, a pre-trained model responsible for detecting and localizing hand joint keypoints in each video frame. The second is a CNN trained by the author, responsible for identifying static and dynamic gestures from a matrix of keypoints organized over time, as explained below.

\subsection{Gesture Detection}

\begin{figure}[H]
    \centering
    \includegraphics[width=0.75\linewidth]{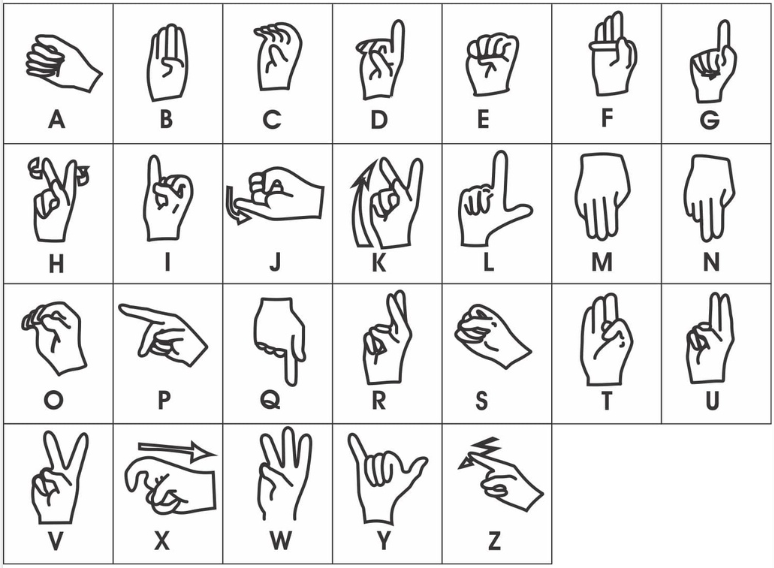}
    \caption{LIBRAS Fingerspelling Alphabet}
    \label{fig:libras}
\end{figure}

This project uses gestures from the LIBRAS alphabet (Língua Brasileira de Sinais — Brazilian Sign Language), the official language of the deaf community in Brazil. It is a visual-gestural language in which communication takes place through signs made with the hands, the body, and facial expressions (Figure~\ref{fig:libras}).

Some letters were associated with devices in a home automation system: "A" for air conditioning, "C" for curtains, "J" for windows, and "L" for lights. The letter J is a dynamic gesture, as is the toggle command (on/off), represented by opening and closing the fingers with the hand in a pear shape pointing toward the camera.

The MediaPipe Hand Landmarker library \cite{IEEEhowto:zhang} is a tool specialized in detecting and localizing hand landmarks in images and videos. It uses a machine learning model trained to identify keypoints such as fingertips, the base of the thumb, and the center of the palm. These landmarks are represented in image coordinates (relative) and, optionally, in world coordinates (absolute).

The library also offers configurations to adjust the maximum number of detected hands and the confidence thresholds for detection and tracking. The results include information about the handedness (left/right) and the detected landmarks, enabling developers to create interactive applications based on hand movements.

\begin{figure}[H]
    \centering
    \includegraphics[width=1\linewidth]{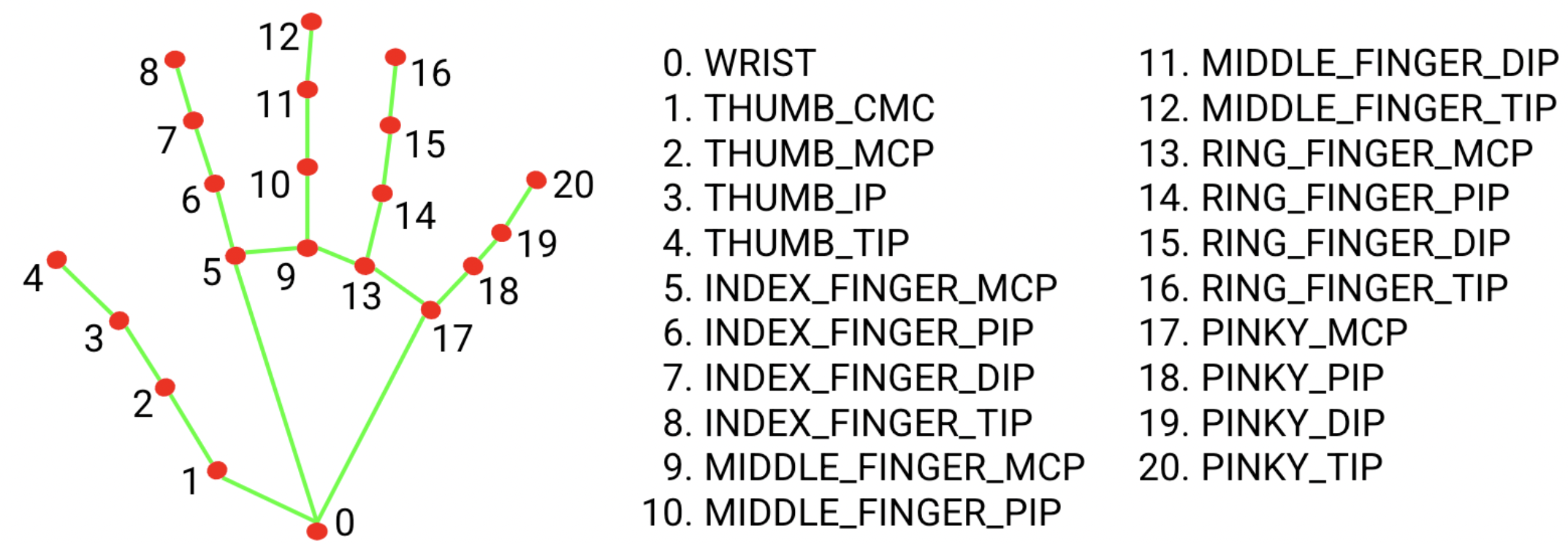}
    \caption{MediaPipe Keypoint Scheme}
    \label{fig:mediapipe_points}
\end{figure}

Figure~\ref{fig:mediapipe_points} shows the keypoint scheme used by the library. For each joint keypoint, a set of coordinates (x, y, z) is provided, allowing its spatial localization, relative or absolute depending on the configuration. In summary, the output includes: hand orientation (left or right), coordinates of the 21 hand keypoints, and the confidence level of keypoint detection.

The \textit{HandLandmarker full} model was adopted, capable of processing images of $192\times192$ or $224\times224$ pixels. For this model variant, the average latency for recognizing a hand position is 17.12\,ms for CPU processing and 12.27\,ms for GPU processing \cite{IEEEhowto:zhang}. This means that several images can be processed in a short time interval, an essential characteristic for gesture detection.

\subsection{Gesture Capture}

The more frames used to capture a movement, the more detail can be extracted. Subtle intermediate movements that help describe the gesture can be easily observed. However, there are limits imposed by the hardware (cameras in particular) and by the processor, CPU or GPU, of the computer.

To this end, experiments were conducted with different gesture expression speeds, varying the number of frames and the capture interval. There is considerable subjectivity in this definition, since LIBRAS-fluent people express gestures more quickly than beginners, who take longer to form a gesture.

In general, the goal was to accommodate gestures within an interval of approximately 1,000\,ms. This interval provides some comfort for adjusting and sustaining the hand position, particularly for dynamic gestures. A rate of 30 frames per second (fps) was achieved. The extra latency relative to the Hand Landmarker model is due to secondary processing required for image treatment and handling of the recognition output.

\begin{figure}[H]
    \centering
    \includegraphics[width=0.6\linewidth]{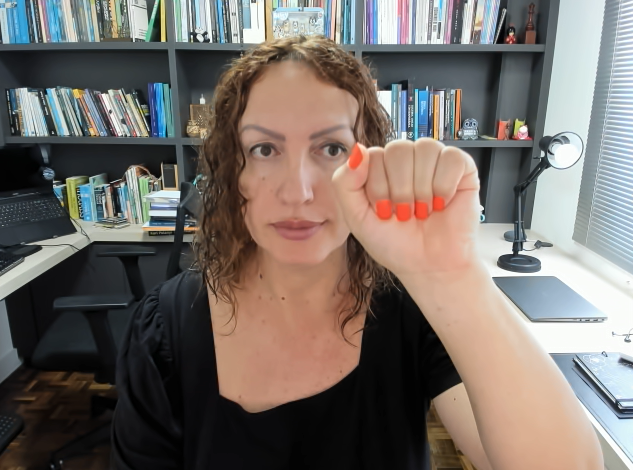}
    \caption{Static Gesture — Letter A}
    \label{fig:capture_A}
\end{figure}

For training the second neural model, 1,254 gestures were captured along with 220 additional samples, captured separately, for validation. Each capture comprises a set of 30 frames. The gestures were divided into 11 classes, covering the following devices and actions: air conditioning, curtains, windows, lights, toggle (on/off), increase intensity, decrease intensity, change light color to blue, change light color to red, change light color to white, and neutral gestures. Figure~\ref{fig:capture_A} illustrates the capture of a static gesture corresponding to the letter A of the LIBRAS alphabet.

The neutral gesture class is important because it allows hands to be present in the scene without causing false detections and the consequent undesired triggering of devices. This class has approximately twice as many captures as the others, since there are various relaxed hand positions that carry no meaning in the context of this study.

To increase model robustness, gestures were expressed at different speeds and with small variations in finger positioning.

\subsection{Motion Matrix}

The keypoints recognized in a set of 30 frames, corresponding to a gesture, were organized into a matrix of dimensions (90$\times$21). The 90 rows correspond to the vertical concatenation of the 30 frames for each of the three coordinate axes: rows 1--30 contain x values, rows 31--60 contain y values, and rows 61--90 contain z values. The 21 columns correspond to the hand joint keypoints. Before inference, each matrix is individually normalized using min-max scaling to the range [0, 255] and converted to a grayscale image, making the model invariant to hand scale and distance from the camera. This organization enables the visualization of static or dynamic gestures through visual patterns.

\begin{figure}[H]
    \centering
    \includegraphics[width=0.2\linewidth]{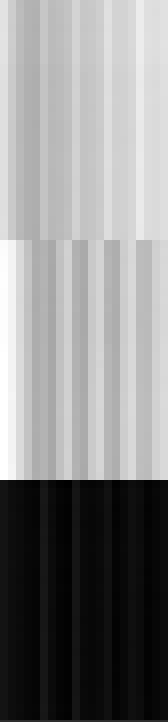}
    \includegraphics[width=0.202\linewidth]{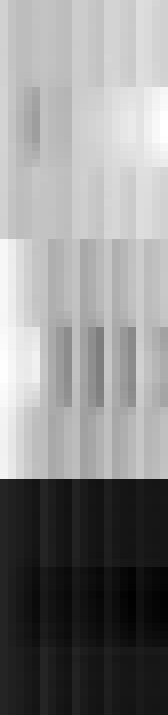}
    \caption{Motion Matrices — "A" and Toggle (On/Off)}
    \label{fig:movement_matrices}
\end{figure}

Figure~\ref{fig:movement_matrices} shows two motion matrices from the dataset. On the left, a static pattern corresponding to the letter A is shown — no variation in coordinate values produces uniform vertical lines across all coordinate sectors. On the right, a dynamic pattern for the toggle (on/off) gesture shows a more complex image, resulting from the variation of joint positions frame by frame.

To facilitate data organization and inspection, the generated matrices were converted into images, as shown in Figure~\ref{fig:movement_matrices}. Captures with evident recording errors, whose patterns were highly discrepant relative to others, were discarded, ensuring a cleaner training set.

\subsection{Model Training}

For the identification of dynamic and static gestures in a motion matrix, a convolutional network was trained. The goal was for the model to recognize gestures with different fluencies and hand position variations. In this regard, convolutional networks offer greater tolerance to translation and scaling of patterns in the image \cite{IEEEhowto:chollet}, which correspond precisely to the time (y) and position (x) dimensions.

\begin{figure}[H]
    \centering
    \includegraphics[width=0.95\linewidth]{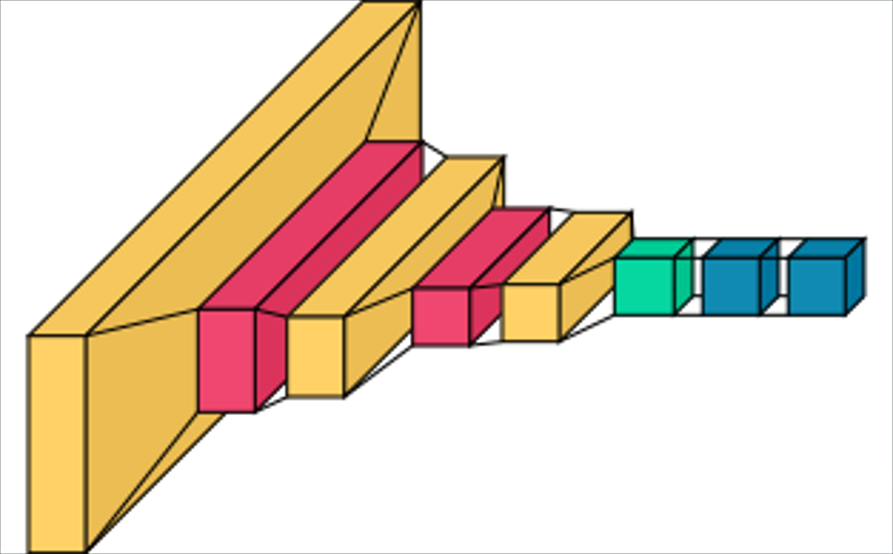}
    \includegraphics[width=0.95\linewidth]{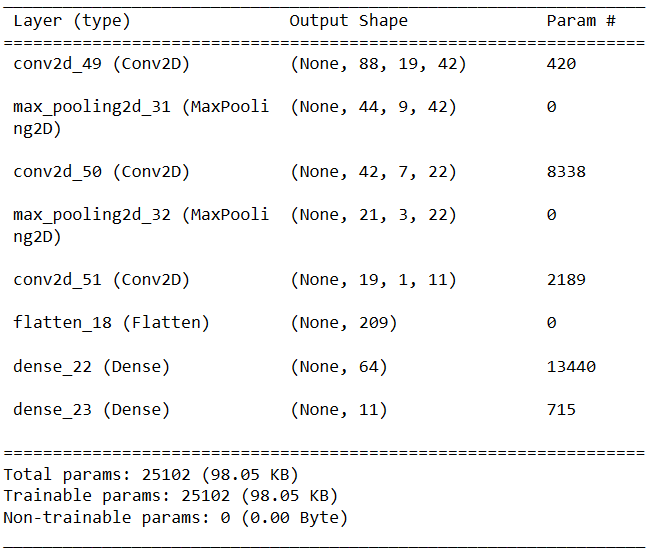}
    \caption{Second Model Architecture — Convolutional Network}
    \label{fig:cnn}
\end{figure}

Figure~\ref{fig:cnn} shows the network architecture, composed of three convolutional layers and two dense layers. The first convolutional layer has 33 filters with a (3$\times$3) kernel, followed by a MaxPooling (2$\times$2) layer. The second and third convolutional layers have 64 filters each, also with a (3$\times$3) kernel. The intermediate dense layer has 64 neurons. The activation function was set to \textit{relu} for all layers except the last, which uses \textit{softmax}. To improve training stability, L1 regularizers ($\lambda$=0.0001) were applied to the second and third convolutional layers and to the intermediate dense layer. The chosen optimizer was Adam, with a learning rate of 0.000001, batch size of 8, and categorical cross-entropy loss \cite{IEEEhowto:chollet}.

Experiments were conducted with different numbers of filters and neurons, with the total parameter count ranging from 2,000 to 253,000. In the smallest configuration, recognition was impaired, with excessive classification uncertainty and consequent performance loss. In the largest configuration, classification was highly precise but computationally expensive — negligible for a modern personal computer processor, but significant for embedded systems with limited memory and processing resources.

The best trade-off between performance and model size was found at around 25,000 parameters. In this configuration, gesture identification was very fast.

\begin{figure}[H]
    \centering
    \includegraphics[width=0.95\linewidth]{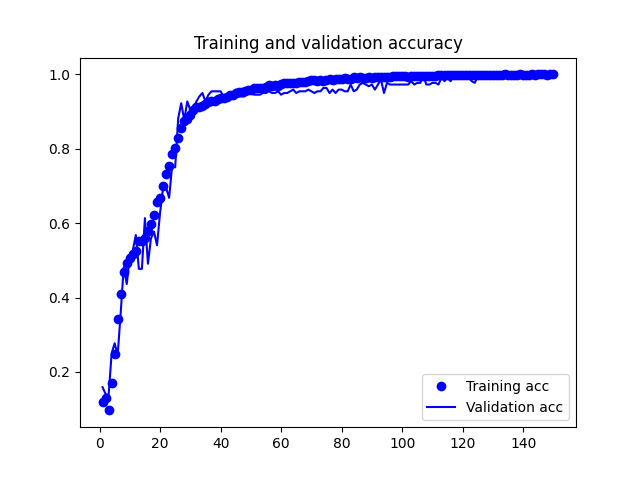}
    \caption{Training Accuracy}
    \label{fig:accuracy}
\end{figure}

Figure~\ref{fig:accuracy} shows the evolution of model accuracy over 150 training epochs. The effect of the regularizers can be observed in the small oscillation present in the graph. In experiments without regularizers, convergence was less smooth and did not stabilize at the end. An important observation is the proximity between training and validation accuracy — both converging to values close to 100\%. This behavior is expected given that both sets were generated by the same person: anatomical and expressive homogeneity eliminates part of the variability that normally causes divergence between training and validation. Although this result indicates good convergence for the available data, it should not be interpreted as a guarantee of generalization to other users. Collecting gestures from a larger group of people would be necessary to evaluate model robustness under more diverse conditions.

\begin{figure}[H]
    \centering
    \includegraphics[width=0.95\linewidth]{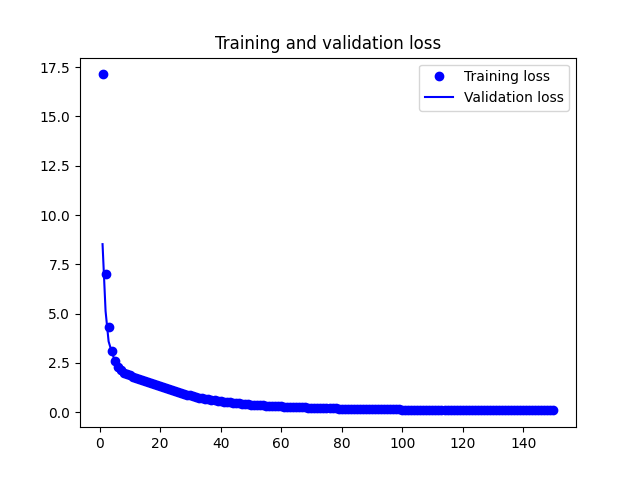}
    \caption{Training Loss}
    \label{fig:loss}
\end{figure}

Similarly, the training loss shown in Figure~\ref{fig:loss} also exhibits fast convergence and low oscillation, which can be attributed to the same factors described above.

In general, the model training yielded good results: both training and validation accuracy reached 100\% on several occasions across the different tested architectures. In the adopted configuration, training accuracy reached 100\% and validation accuracy reached 99.55\%.

\subsection{Dynamic Gesture Detection}

Once the model is trained, an inference heuristic must be established. The initial idea would be to capture a set of 30 images per second and attempt to identify gestures within those intervals. However, this approach has a drawback: a gesture may be partially captured, with part of it in one block and the remainder in the next. Dynamic gestures would thus be easily missed.

\begin{figure}[H]
    \centering
    \includegraphics[width=0.75\linewidth]{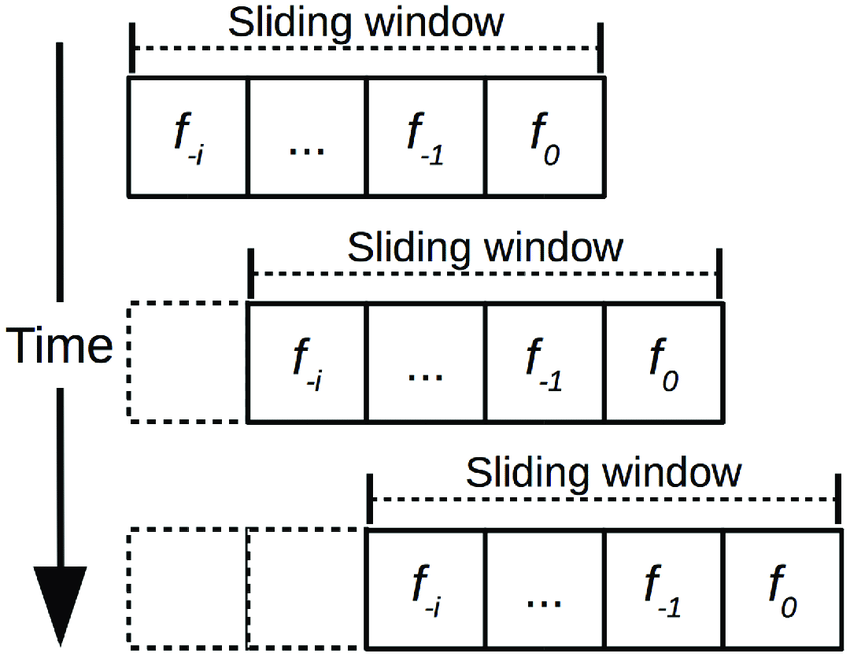}
    \caption{Sliding Window}
    \label{fig:window}
\end{figure}

One way to address this issue is through a sliding window for frame capture, as illustrated in Figure~\ref{fig:window}. In this method, each new frame captured by the camera is inserted $n$ times into the buffer — where $n$ is a configurable speed parameter. In the adopted implementation, $n=3$: each real frame is triplicated in the buffer, so a window of 30 entries corresponds to approximately 10 real frames. Each time the buffer reaches 30 entries, the matrix is assembled and submitted to the model. Only predictions with confidence above 98\% are considered valid. With each new inserted frame, the 3 oldest entries are removed from the front of the buffer, ensuring continuous overlap between consecutive windows. When a gesture is recognized and executed, the buffer is fully cleared to prevent repeated detections of the same gesture. If hands leave the scene, the buffer is also cleared immediately.

This technique offers some advantages: by triplicating frames in the buffer, gestures can be expressed more quickly and fluently, at the cost of a slight reduction in the temporal granularity of the analysis.

On the other hand, the technique also has limitations. Gestures whose movements are a subset of the movements of other gestures tend to be recognized prematurely. One example is the letter I, whose movement is contained within the letter J: when performing J, the system may recognize I before the gesture is completed. For such cases, it is recommended to assume a neutral position or remove the hands from the scene between gestures, while also observing the appropriate cadence.

From a practical standpoint, this cadence adjustment is similar to learning any human-machine interface. For most home automation sequences tested — such as changing context, activating a device, increasing intensity, and changing color — it was not necessary to assume neutral postures or remove the hand from the scene, as the gestures involved do not present movement overlap.

\section{Simulations and Results}

To support the simulation process, a simple operating interface was created. It is capable of capturing gestures and presenting visual feedback of the actions requested by the user. Tests were conducted at different capture speeds, under two lighting conditions — low and normal —, with variations in gesture expression, and across different action sequences. Tests were carried out on a personal computer equipped with an Intel Core i7 processor, 16\,GB of RAM, and a Logitech Brio camera, without a dedicated GPU for inference.

\begin{figure}[H]
    \centering
    \includegraphics[width=0.75\linewidth]{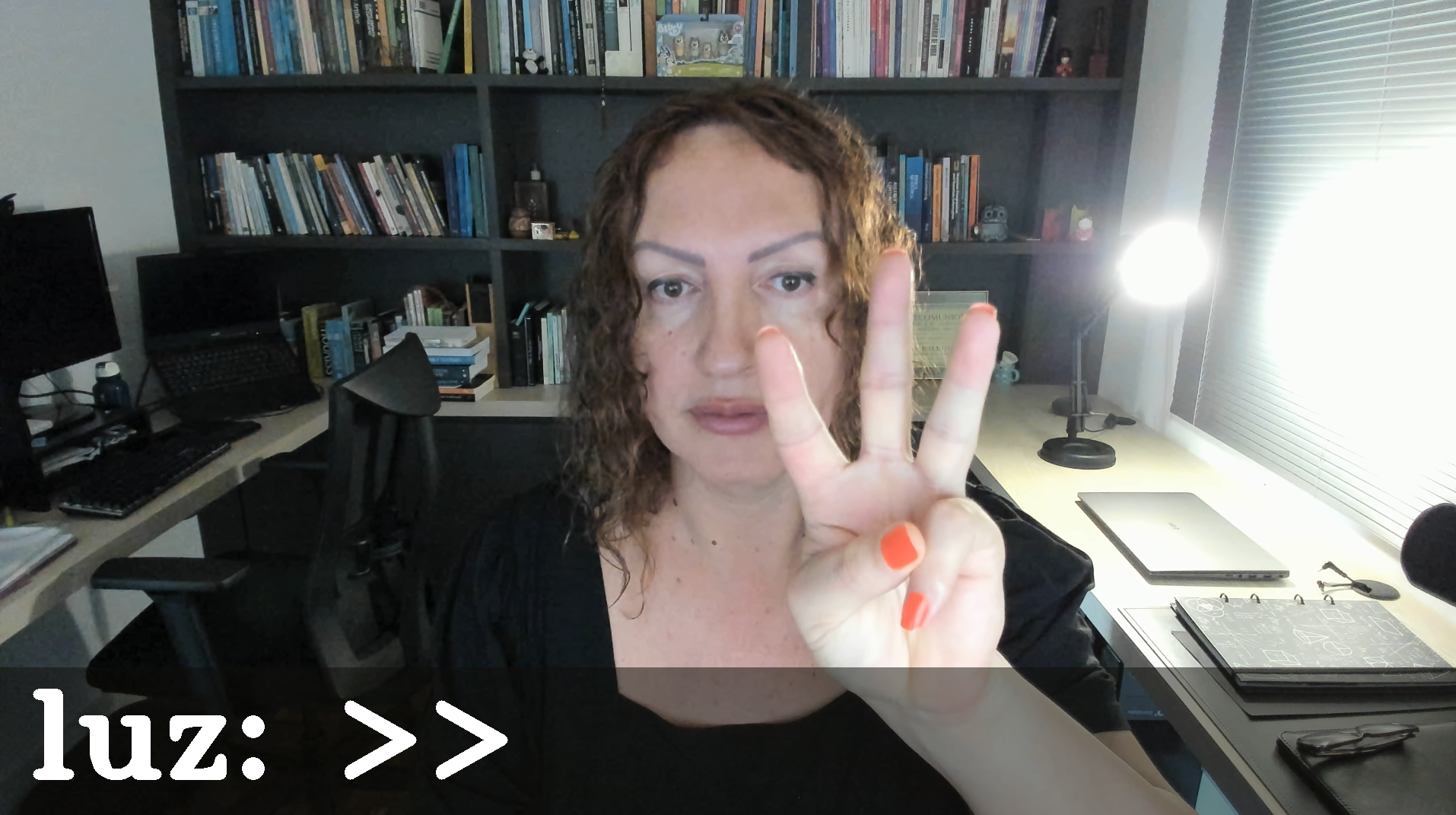}
    \caption{Gesture Detection under Low-Light Conditions}
    \label{fig:prod1}
\end{figure}

Figure~\ref{fig:prod1} shows the execution of the command to change the light color to white. The text color was also changed to white, indicating that the action was executed. A test sequence was conducted in a low-light environment and achieved high accuracy. However, the results are directly related to the quality of the camera used, whose sensor is capable of capturing images even in low-light conditions.

Table~\ref{tab:resultados} summarizes the results obtained under the two tested lighting conditions. For each condition, 100 consecutive recognitions of isolated gestures were performed, with buffer reset between each attempt. Gestures were expressed with variations, aiming to reach recognition limits without excessive distortion.

\begin{table}[H]
\caption{Test Results by Lighting Condition}
\label{tab:resultados}
\centering
\renewcommand{\arraystretch}{1.3}
\begin{tabular}{lcc}
\hline
\textbf{Condition} & \textbf{Correct} & \textbf{Main error} \\
\hline
Low light & 95/100 & C as neutral \\
Normal light & 92/100 & White (5), C (3) \\
\hline
\end{tabular}
\end{table}

The full sequence used in the normal-light test included context switching, device activation, intensity increase, color change to white, intensity decrease, and deactivation. In both conditions, the letter C was the gesture with the highest error rate, being confused with relaxed hand positions.

\begin{figure}[H]
    \centering
    \includegraphics[width=0.75\linewidth]{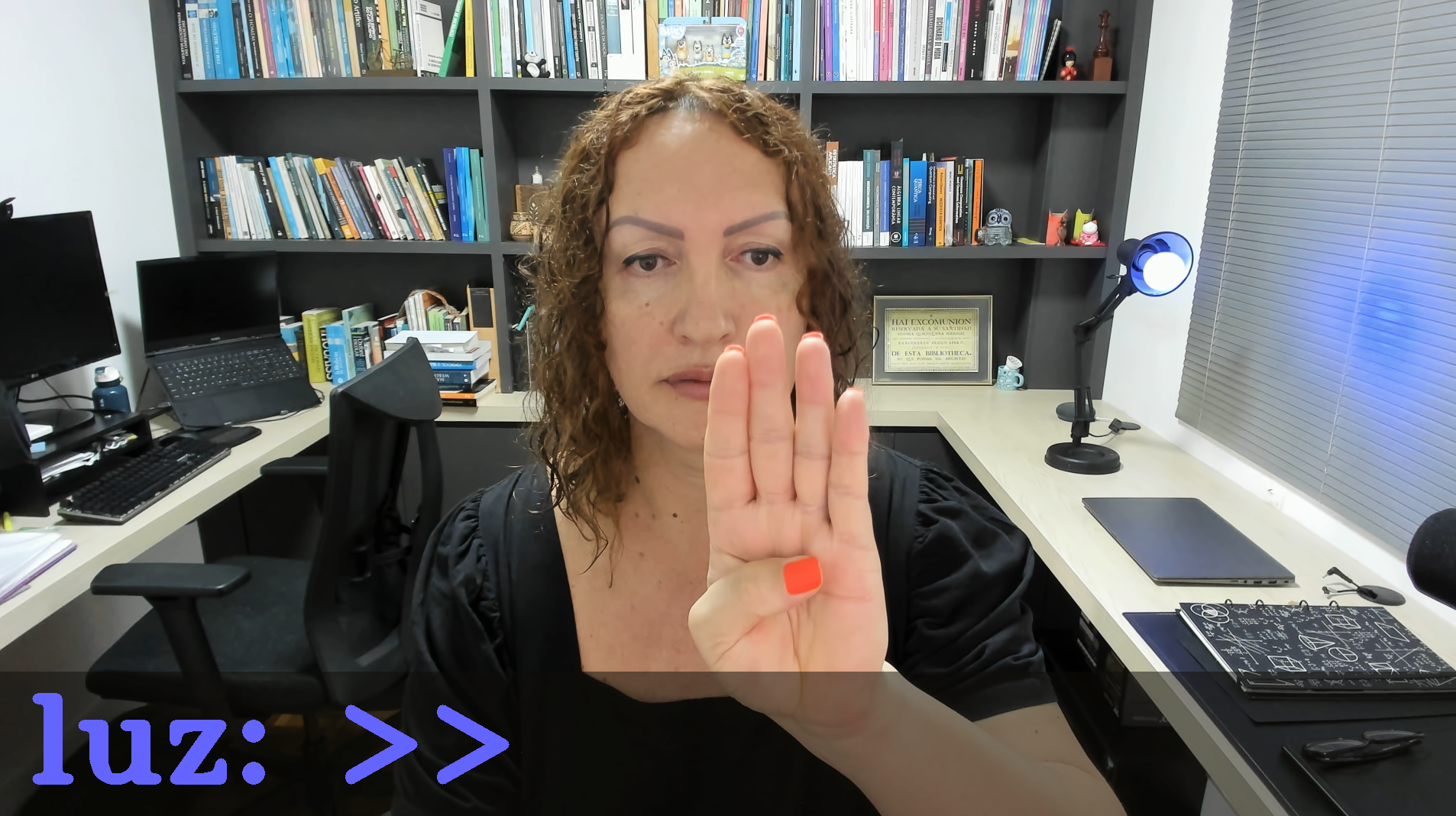}
    \caption{Gesture Detection under Normal Lighting}
    \label{fig:prod2}
\end{figure}

Although the test set is limited, it highlights the need to increase the diversity of movements and anatomical features during gesture capture. More captures with rightward and leftward tilt and variations in finger positioning — such as the thumb pointing up, down, or at an angle, as illustrated in Figure~\ref{fig:prod2} — would help ensure good accuracy across different users.

The results obtained are comparable to those reported in the literature for similar scenarios: Alabdullah et al. \cite{IEEEhowto:alabdullah} achieved 92.57\% in gesture-based home control, while Alves et al. \cite{IEEEhowto:alves}, using a conceptually similar approach with skeletal images and CNN, obtained 93\% on the MINDS-Libras dataset with 12 distinct users. It is worth noting that the tests presented here were conducted with a single user under low and normal lighting conditions, which favors accuracy but limits the evaluation of model generalization.

The need for systematic tests to evaluate the system more rigorously should be emphasized. Ideally, a formal test set independent from the validation set should be employed, properly separating development and evaluation phases. Additionally, experiments with multiple users would be needed, analyzing the influence of personal characteristics — such as expression style, rhythm, and anatomical variations — on the accuracy of isolated or sequential gesture detection.

\section{Conclusion}

This paper presented a method for dynamic LIBRAS gesture recognition applied to device control in a home automation system. The three proposed contributions were demonstrated throughout the work: (1) the spatiotemporal matrix representation (90$\times$21) of skeletal keypoints extracted by MediaPipe proved effective for capturing visual patterns of static and dynamic gestures, enabling visual inspection and filtering of training data; (2) the sliding window strategy with temporal frame triplication ($n=3$) enabled continuous real-time inference without recurrent networks, keeping the architecture simple ($\sim$25,000 parameters) and compatible with general-purpose hardware; and (3) validation in a home automation scenario with 11 LIBRAS gesture classes achieved 95\% accuracy under low-light conditions and 92\% under normal lighting, results comparable to those reported in the literature for similar systems \cite{IEEEhowto:alabdullah}\cite{IEEEhowto:alves}.

Compared to approaches that use finite state machines \cite{IEEEhowto:hakim} or recurrent models \cite{IEEEhowto:ridwang} to model temporal sequences, the proposed architecture dispenses with additional context control components, reducing implementation complexity and computational cost.

For more rigorous evaluation, it is necessary to expand the training dataset with gestures captured from multiple users with different anatomical characteristics. Generalization assessment should be conducted with a formal, independent test set under varied lighting and hardware conditions.

As future work, we intend to expand the dataset with gestures from multiple users, incorporate variable lighting as an explicit training variable, and evaluate the portability of the system on low-cost embedded hardware. Extending the gesture vocabulary beyond the current LIBRAS subset also represents a relevant direction to increase the practical applicability of the system.

\end{document}